\documentclass[twocolumn]{svjour3}  
\usepackage{amsfonts,amsmath,amssymb}
\usepackage{epsfig}
\usepackage{listings}
\usepackage{pdfpages}
\usepackage{float}
\usepackage{url}
\usepackage{comment}
\usepackage{textcomp}
\usepackage{xcolor}
\usepackage{latexsym}
\usepackage{graphicx}
\usepackage{hyperref}
\usepackage{soul}
\usepackage{picins}

\definecolor{mygreen}{rgb}{0,0.6,0}
\definecolor{mygray}{rgb}{0.5,0.5,0.5}
\definecolor{mymauve}{rgb}{0.58,0,0.82}
\usepackage{lipsum}
\usepackage[noadjust]{cite}

\begin{document}

\title{Real-Time Traffic End-of-Queue Detection and Tracking in UAV Video}
\author{Russ Messenger \and Md Zobaer Islam$^*$ \and Matthew Whitlock \and Erik Spong \and Nate Morton \and Layne Claggett \and Chris Matthews \and Jordan Fox \and Leland Palmer \and Dane C. Johnson \and John F. O'Hara \and Christopher J. Crick \and Jamey D. Jacob \and Sabit Ekin$^*$
\thanks{*Corresponding authors:\\
Md Zobaer Islam (215 General Academic Building, OSU, Stillwater, OK 74078, USA, zobaer.islam@okstate.edu)\\[4pt]
Sabit Ekin (Fermier Hall, 3367 TAMU, 466 Ross St, College Station, TX 77843, USA, sabitekin@tamu.edu)}
}

\authorrunning{R. Messenger et al.}

\institute{
R. Messenger, M. Z. Islam, M. Whitlock,  N. Morton, C. Matthews, J. Fox, L. Palmer, J. O’Hara and S. Ekin
\at School of Electrical and Computer Engineering, Oklahoma State University, Stillwater, OK, USA \\
\email{russ.messenger, zobaer.islam, matthew.whitlock, nathanial.morton, cmatt11, jordan.fox, lelanep, oharaj, sabit.ekin(@okstate.edu)}
\and
D. C. Johnson
\at Unmanned Systems Research Institute (USRI), Oklahoma State University, Stillwater, OK, USA \\
\email{dane.johnson@okstate.edu} 
\and
C. J. Crick
\at Department of Computer Science, Oklahoma State University, Stillwater, OK, USA \\
\email{chriscrick@cs.okstate.edu}
\and
E. Spong, L. Claggett, and J. D. Jacob
\at School of Mechanical and Aerospace Engineering, Oklahoma State University, Stillwater, OK, USA \\
\email{erik.spong, layne.claggett, jdjacob(@okstate.edu)}
\and
Sabit Ekin
\at Department of Engineering Technology and Industrial Distribution, Texas A\&M University, College Station, TX, USA \\
\email{sabitekin@tamu.edu} 
}

\date{Accepted: October 6, 2023}
\maketitle

\begin{abstract}
Highway work zones are susceptible to undue accumulation of motorized vehicles which calls for dynamic work zone warning signs to prevent accidents. The work zone signs are placed according to the location of the end-of-queue of vehicles which usually changes rapidly. The detection of moving objects in video captured by Unmanned Aerial Vehicles (UAV) has been extensively researched so far, and is used in a wide array of applications including traffic monitoring. Unlike the fixed traffic cameras, UAVs can be used to monitor the traffic at work zones in real-time and also in a more cost-effective way. This study presents a method as a proof of concept for detecting End-of-Queue (EOQ) of traffic by processing the real-time video footage of a highway work zone captured by UAV. EOQ is detected in the video by image processing which includes background subtraction and blob detection methods. This dynamic localization of EOQ of vehicles will enable faster and more accurate relocation of work zone warning signs for drivers and thus will reduce work zone fatalities. The method can be applied to detect EOQ of vehicles and notify drivers in any other roads or intersections too where vehicles are rapidly accumulating due to special events, traffic jams, construction, or accidents.
\end{abstract}

\section{Introduction}
The overall purpose of this study is to develop an efficient, reliable, and safe Intelligent Traffic System (ITS) to assist in traffic data collection and dissemination for smart work zone traffic management using low-cost Unmanned Aerial Vehicles (UAV). The proposed framework followed a holistic approach to provide a comprehensive guideline, including regulations and policies for the use of UAV, for an efficient real-time processing and analysis of a UAV-based traffic monitoring and management system.

\begin{figure*}[ht]
  \centering
  \includegraphics[width=\textwidth, height=1.98in]{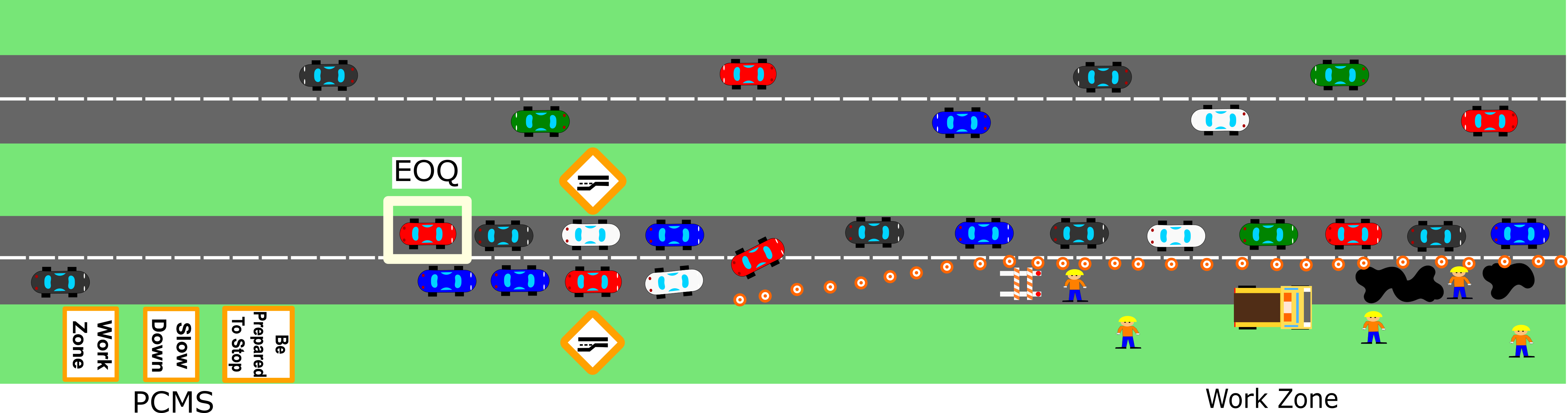}
  \caption{Illustration of a work zone in highway.}
  \label{img:HighwayWorkZoneV1}
\end{figure*}

Highway work zones are defined as a place where roadwork is taking place, but it does not block the whole roadway. The roadwork will cause only partial blockage by lane closure. Thus, vehicles will be slowed down near work zones but still can pass the work zone through the other available lanes. Fig.~\ref{img:HighwayWorkZoneV1} depicts an example of such work zones. Highway work zones present elevated safety risks to drivers, passengers, and construction workers. In 2017, the number of fatal road crashes due to work zone was increased from that of 2016 by 3\% in USA. In that year, 799 people were killed in various work zone crashes~\cite{nhtsa_2016}. The most common types of work zone crashes were rear-end crashes, running into slow moving or stopped vehicles, and crashes involving fixed objects such as barriers or equipment when drivers try to avoid the rear-end crashes. The majority of fatal work zone crashes occurred on roads with speed limits exceeding 50 mph, particularly urban freeways and arterials~\cite{nhtsa_2016}. When the number of vehicles attempting to enter the work zone approaches or exceeds the capacity of the work zone, queues of stopped or slowed vehicles build up rapidly upstream. Therefore, there is a need for the dynamic management of work zone traffic signs using ITS technologies. Examples of ITS applications include real-time traffic monitoring, real-time queue warning using advance variable message signs, variable speed limits upstream of the work zone to manage the speed of approaching traffic, and incident management.

The use of technology for traffic management has been implemented in a variety of ways, but often requires installation of expensive, specialized, and location-limited hardware. With improvements in video and image processing technologies, it has become possible to detect and track vehicles using only video footage~\cite{jazayeri_cai_zheng_tuceryan_2011}. These advancements spurred investigation into analyzing traffic trends and managing infrastructure using only pre-existing traffic cameras~\cite{yang_hou_jhang_2013}. However, these analyses are limited to regions with pre-existing traffic cameras.

\begin{figure*}[ht]
  \centering
  \includegraphics[width=0.82\textwidth]{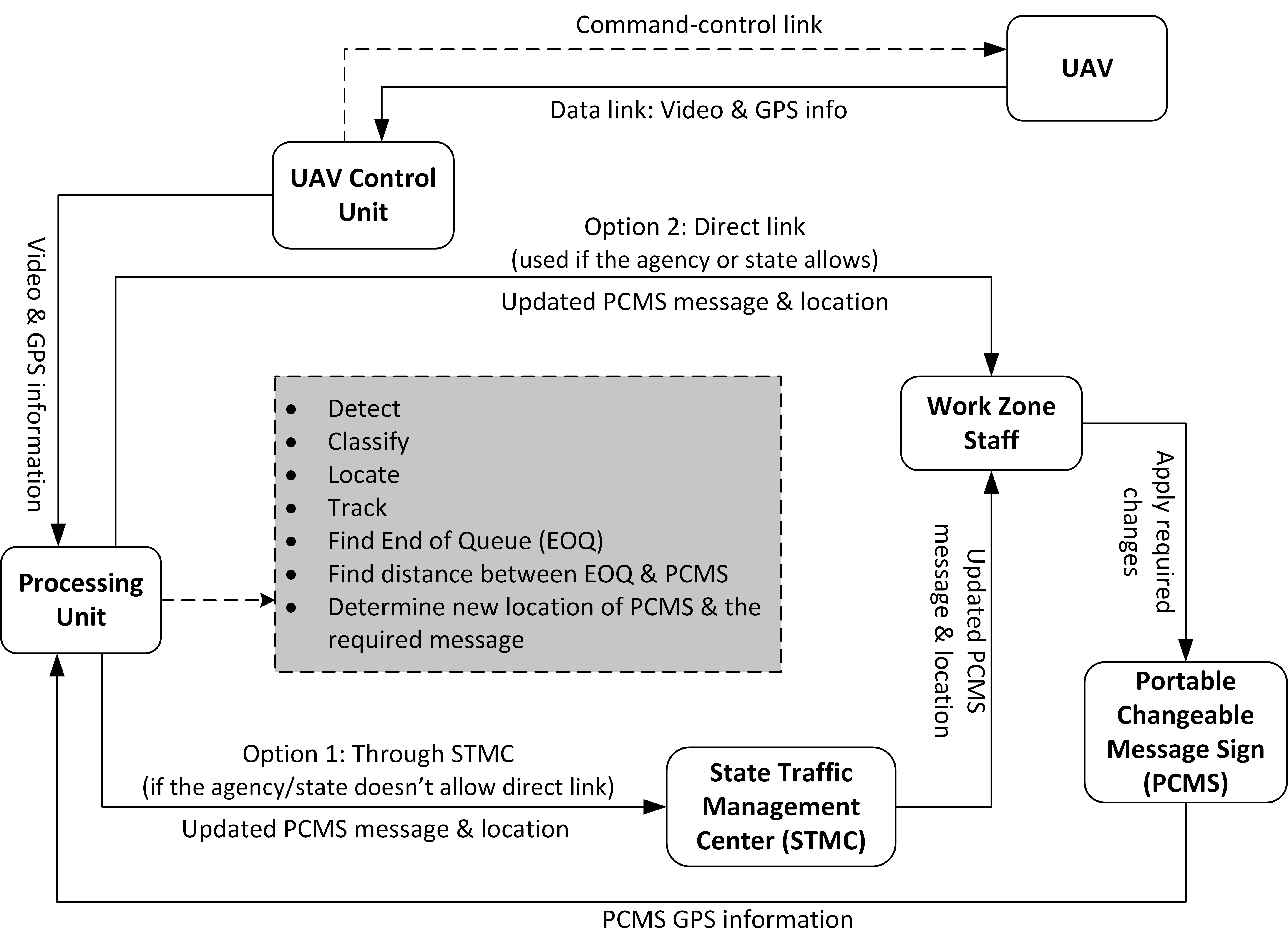}
  \caption{Illustration of framework for UAV-assisted work zone traffic management.}
  \label{img:fig_1_corr}
\end{figure*}

UAVs have been increasingly popular and widely used for surveillance and reconnaissance. They have great potential as a part of ITS technologies for providing fast and real-time aerial video images in both visible and multi-spectral wavelengths, such as infrared. They require less costly infra-structure and can provide a more agile response than stationary traffic cameras and Manned Aerial Vehicles (MAV) in traffic-flow monitoring, especially at the time of road-construction, traffic jam or any other emergency situations~\cite{Barmpounakis_vlahogianni_golias_2016}.  Efficiently and accurately detecting and tracking moving objects in UAV video has been the subject of research for many years, and is still actively being researched on~\cite{Kanistrtas_Martins_Rutherford_2013, Wang_Liu_Zhou_2017, multi_tracking_aerial, li_ye_chung_kolsch_wachs_bouman_2016}. Fast detection of vehicles from images captured by UAVs has been researched, and an algorithm for doing this was developed in~\cite{gleason_nefian_bouy_fong_bebis_2011}. In numerous prior research works, video footage from UAVs are used to detect multiple moving targets, track their velocities and trajectories~\cite{Wang_Liu_Zhou_2017, multi_tracking_aerial, li_ye_chung_kolsch_wachs_bouman_2016} and also geo-reference them~\cite{baykara_biyik_gul_onural_ozturk_yildiz_2017, Collins_Barooah_Hespanha_2016, Zhou_2010}. In more recent efforts, UAVs have been utilized in road traffic monitoring~\cite{khan_ectors_bellemans_janssens_wets_2017, wang_chen_yin_2016, huang_chen_yang_cheng_2017, Salvo_Caruso_Scordo_2014, Hosseinpoor_Samadzadegan_Dadrasjavan_2016} and video processing frameworks using image processing~\cite{Khan_Ectors_bellemans_Wets_2017, Khan_Ectors_bellemans_Wets_2018}, convolutional neural networks~\cite{baykara_biyik_gul_onural_ozturk_yildiz_2017} and Kalman filtering~\cite{Khan_Ectors_bellemans_Wets_2018, baykara_biyik_gul_onural_ozturk_yildiz_2017, multi_tracking_aerial, li_ye_chung_kolsch_wachs_bouman_2016, bruin_booysen_2015, Dille_Grocholsky_Singh_2011, Hosseinpoor_Samadzadegan_Dadrasjavan_2016} for detecting and tracking moving vehicles have been proposed for this.

One of the notable applications of ITS found in literature is determination of queue length or queue tail detection. The end-of-queue (EOQ) in this paper shall be defined as the starting point where traffic has to significantly travel slower than the posted speed limit. Various qualitative and quantitative approaches including image processing of CCTV camera images~\cite{Hoose2016} and videos~\cite{Zanin2003}, corner detection~\cite{Albiol2011}, edge detection~\cite{Cai2010}, motion detection and vehicle detection~\cite{Fathy1995, AlOkaishi2019} have been proposed for vehicle queue detection and queue length determination in highways and intersections during roadworks, special events, high traffic or red signal light scenarios. These approaches further the drive for traffic management by using technology in which road signs may be relocated based on the location of the end-of-queue of vehicles. Such relocation of road signs are possible using Portable Changeable Message Signs (PCMS) available in the market~\cite{pcms1,pcms2}. Vehicle queue length or end-of-queue information can be utilized to prevent accidents and minimize traffic delays by determining new speed limits, displayed via PCMS, calculating waiting times for vehicles~\cite{Shirazi2015}, and notifying drivers via either in-vehicle message~\cite{Craig2017} or messaging signage regarding slow moving traffic ahead.

In this paper, we develop a method of detecting EOQ of the upstream vehicle traffic and demonstrate a proof of concept. The framework is shown in Fig.~\ref{img:fig_1_corr}. We used a UAV to capture the video of traffic. The UAV transmitted the video and its Global Positioning System (GPS) information wirelessly to the UAV controller unit and then to the video processing unit. The processing unit utilized existing techniques of image/signal processing to analyse the video, track the vehicles and find the EOQ of the traffic. The real-time traffic information would be transmitted to workers, traffic control personnel, and variable message signs.

In the next section (Section~\ref{hardware}), current UAV regulations and hardware details used for various units of the proposed framework are discussed. In Section~\ref{software}, the developed software is presented in details including the algorithm for end of-queue detection. Finally, the testing conditions and test results are outlined in Section~\ref{testing}.

\section{UAV Technology and Hardware}
\label{hardware}
In order to thoroughly test the software, a UAV was selected that would make a steady platform to take video as well as meet safety and regulatory requirements. All communication between the hardware used either cellular or WiFi wireless radios, unless stated otherwise.

\subsection{UAV Regulations}
The main regulatory body that will influence and restrict the UAV design is the Federal Aviation Administration (FAA). The UAV will fall under Part 107, rules for small unmanned aircraft weighting less than 55 pounds, of FAA regulations~\cite{droneregulations}. Below are the major UAV design and operating restrictions:

\begin{itemize}
  \item Operator must keep the UAV within their unaided line of sight.
  \item UAV can operate 30 minutes before official sunrise to 30 minutes after sunset local time.
  \item Minimum weather visibility of three miles.
  \item Maximum altitude of 400 feet above the ground.
  \item Maximum speed of 100\,mph.
  \item UAV may not fly over anyone not directly participating in the UAV operation.
  \item UAV may not fly under a covered structure.
  \item UAV may not fly inside a covered stationary vehicle.
  \item Operator may not be flying the UAV from a moving vehicle.
  \item External load is allowed if it is securely attached and does not adversely affect the flight characteristics or controllability of the aircraft.
  \item The attached systems, payload, and cargo must weigh less than 55 pounds in total.
  \item Operation in Class G airspace is allowed without Air Traffic Control (ATC) permission.
  \item Operation in Class B, C, D, and E non-surface airspace needs ATC approval.
\end{itemize}

These regulations did limit the number of suitable locations for initial testing, but did not affect our software or hardware choices. UAV operations are restricted to Class E non-surface and Class G airspace. This was not an issue as the large majority of active airspace is Class E and Class G.

\subsection{UAV Hardware}
The DJI Matrice 600 (M600)~\cite{dronewebsite} was selected due to its availability, safety features, and versatility.  There were many sensors built into the UAV; however, the altitude sensor was replaced for better accuracy.  The built-in barometric pressure altitude sensor could only measure altitude relative from the takeoff position, so the LightWare LW20/C~\cite{lidarwebsite} lidar was selected. The lidar had an accuracy \textpm10 cm, a resolution of 1 cm, and a range of 100 m. This altitude measurement was used in the calculations for detecting the GPS location of the queue of vehicles.

A Raspberry Pi 3B was used to add extra sensors to the DJI M600.  It was responsible for getting data from the UAV’s flight controller and the added lidar. The Raspberry Pi was powered by the UAV, and it used the DJI's onboard Software Development Kit (SDK) to interface with the UAV. \label{microcontroller}

The DJI X3 camera was used to capture video footage from the UAV. It is capable of taking footage in 4K, but for the processing of the end-of-queue, it will have to be downscaled to 720p. The X3 camera has built-in stabilization, making it easier to detect the movement of vehicles using image processing.  The built-in DJI software would also automatically send the camera footage to the ground through its control radio and thus simplifying communications.

The Raspberry Pi was unable to use the UAV's built-in control radio to send data to the ground, so a telemetry radio was also added. To accomplish this task, a RFD 900+~\cite{rfd900website} was used. Since the UAV's high-power radio could interfere with any radio that used the 2.4 or 5\,GHz bands, 900\,MHz band of the RFD 900+ is used to communicate. An FTDI cable was plugged into the Pi's USB port and into the radio. The radios were set up using the software provided by the manufacturer.  The main change from the default setting was the baud rate. It was set to 9,600 instead of 57,600 for improved bit error rate. Another RFD 900+ was plugged into a laptop on the ground that was responsible for all of the processing. The RFD 900+ behaves exactly like a serial port in the program.

\subsection{Processing Unit}

A laptop with an Intel i7-8750H 2.2-4.2 GHz processor was chosen for sufficiently fast image processing. The laptop was also equipped with a Nvidia GTX 1060 GPU and an SSD/HDD combo for fast disk reads.

\section{Software Development}
Image processing was used to detect the EOQ from the video captured by the UAV. To summarize, background subtraction was used to remove non-moving objects and highlight moving objects. Then the image is passed through a convolutional kernel to expand the sizes of each pixel of the highlighted objects to remove gaps in a single vehicle due to imperfections in the background subtraction. After that, object detection is done by using a color filter to detect blobs or groups of highlighted pixels. Next, the blobs are tested to determine if they are a vehicle or not. The GPS co-ordinate is then calculated for the leftmost vehicle, which denotes the EOQ.\label{software}

\begin{figure*}[htbp!]
  \centering
  \includegraphics[width=\textwidth]{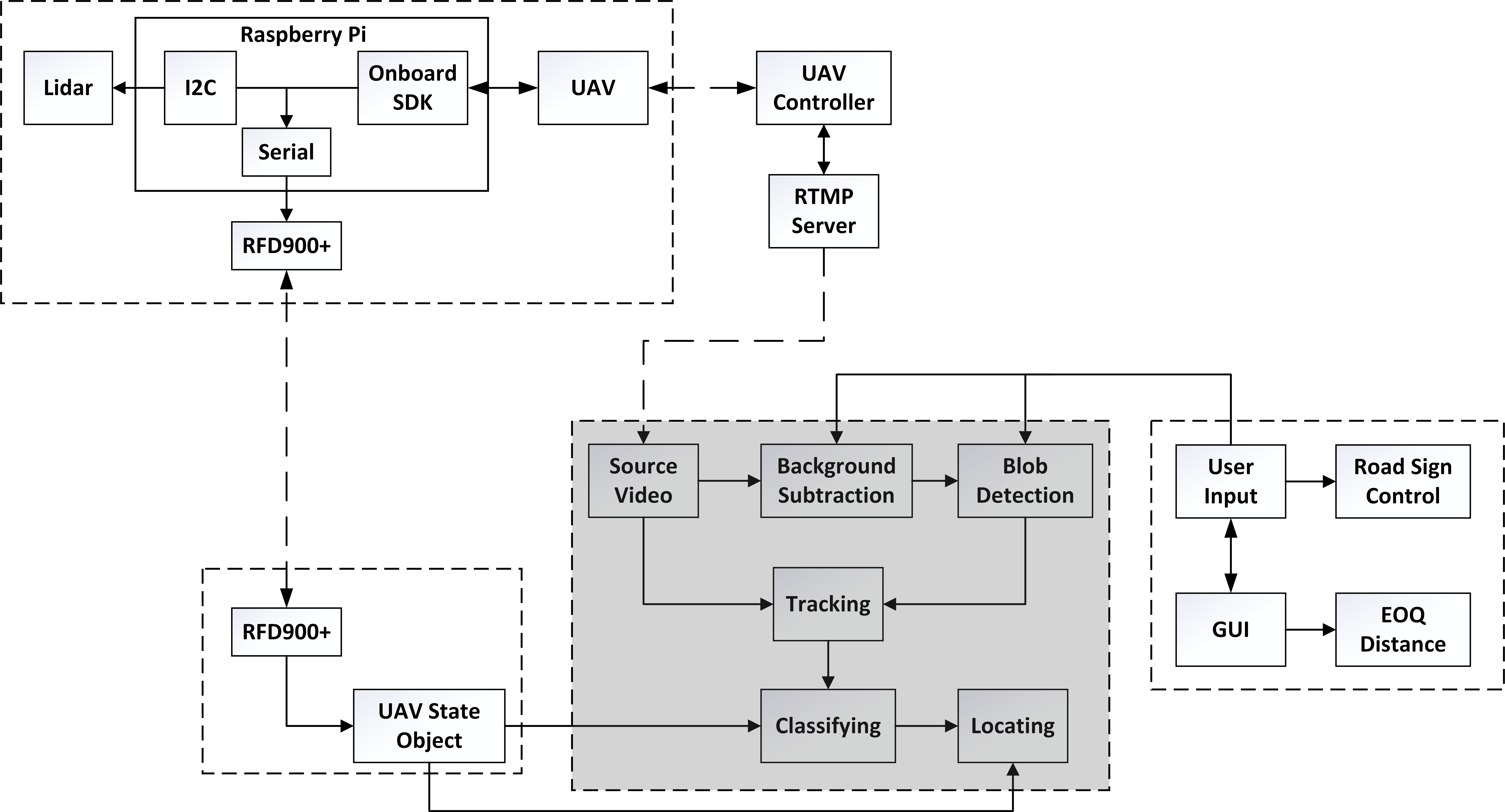}
  \caption{The block diagram of the information flow of the software. Dotted-line regions imply sequential-operation semantics (i.e., a potential bottleneck region).}
  \label{img:Software_flow}
\end{figure*}

With the video and UAV telemetry data, the program follows the computer vision and signal processing steps detailed in the dark box in Fig.~\ref{img:fig_1_corr}.  It is assumed that vehicles are not at a complete standstill for the regions of interest, so background subtraction and contour detection methods were utilized to identify moving objects. Following this approach, each detected object was assigned a tracker, which consists of a class containing some metadata (most notably, past positions of the objects). The trackers used meanshifting and attempted to determine a pattern of movement. Finally, detections were separated into vehicles and noise by applying speed and direction filtering. The data flow through the software is shown in Fig.~\ref{img:Software_flow}. 

The primary bottleneck for computation is with the shaded region of Fig.~\ref{img:Software_flow}. This region operates on the entire frame to track and classify each object sequentially. While it is possible to parallelize this at the object tracking, classifying, and locating level, the software was able to process the streamed 24 frames per second on 720x406 video without the extra effort in real time.

\subsection{Image Processing}

\subsubsection{How it works}
Image processing detects objects, like cars, by going through different algorithmic steps. The first step is to remove the background or stagnant parts from the image, which is called background subtraction. The output consists of white pixels that represent everything moving or non-stagnant in the image. The next step of background subtraction is blob detection. Blob detection scans the image for groups of white pixels or blobs, and depending on the size of the blob, it may be marked as an object.

\subsubsection{Cons}
Though image processing was the chosen method, it did come at the cost of noisier, less robust detection. Image processing (in the method chosen for this work) is unable to detect stationary vehicles and only detects moving objects. This means that some thought needs to be put into the worth of knowing where stationary cars are located, and ways to filter out noise.

\subsubsection{Pros}
The EOQ, in heavy traffic, should always be made up of moving cars. If the queue is moving, then all the cars will be moving.  Even if the traffic stops, the EOQ will still be moving since cars at the EOQ will be slowing down to meet the stopped traffic. Along with this, we considered image processing to be a relatively simpler method compared to the other possible ways of detecting cars in traffic (such as a Convolutional Neural Network or CNN)
. Also image processing was a fast method for analyzing the footage even without high parallelization. This gave  significant room for performance gains, if needed.

\subsection{UAV Telemetry}
In order to find the real-world GPS positions of the vehicles, as well as to calculate accurate speeds given the angle of the camera, the software needs telemetry data from the UAV as well as the camera’s gimbal. The data needed includes UAV's GPS, height, camera yaw and camera pitch.

As discussed in Section~\ref{microcontroller}, the data was pulled from the UAV's flight controller to an onboard Raspberry Pi. The microprocessor sends serial data using the RFD900+ on the 900\,MHz frequency band. The software is designed to function with varying sampling rates, even with varying sampling rates between sensors. To achieve this, the protocol for sending data includes a small three letter header to indicate what telemetry data is being sent. This adds a small overhead to the time required to send the information compared to using an order-based protocol, but it allows data to be sampled at whatever rate the microprocessor is able to manage with each sensor.

\subsection{Background Subtraction}
Background subtraction involves analyzing the frames of a video and determining what parts of the frame are stagnant (the background) and what parts are moving (the objects). It takes constant frames of the UAV's video feed as an input and subtracts from the image the parts that are stationary. It outputs a black and white video feed (with white being the foreground).

There are several ways of determining the background. One way is to assume that the first frame is the background without any objects. Another option is to take an average of all the frames of a video and consider any major deviation as an object. A third method uses a Mixture of Gaussian (MOG) approach. In this method, pixels are represented by a mixture of Gaussian distributions. Each Gaussian distribution is represented by a color and is weighted by the amount of time the pixel remains a certain color~\cite{MOGref}.

The first assumes a stagnant background and that there are no moving objects in the first frame. This method has its disadvantages, since it requires stagnant lighting and it may be challenging to obtain a picture of the background without any moving objects in the image.  The second assumes that the background will change in small amounts over time and that moving objects are rare. Though this method does handle subtle changes in lighting, it fails if moving objects are too frequent. Finally, the last method of doing background subtraction should be able to handle constant moving objects in the image and handle slight movements caused by the instability of the UAV~\cite{MOGref}.

It can be assumed that there will not be the ability to obtain an image of the background by itself and that the images we get will always contain moving objects i.e., vehicles. The lighting of the image is assumed to not change drastically since the UAV will not maintain its position in the air for longer than an hour. With these key details in mind, method 3 was selected for background subtraction (though a combination of method 3 and 2 could be something to consider for subtle lighting changes). For the Image Processing of the UAV's video feed, OpenCV~\cite{opencv} was used for its many libraries on background subtraction. This software can point out/remove the shadows created by objects in the image which could be useful for ensuring that the objects on the image do not appear larger than they really are.

\begin{figure*}[htbp!]
  \centering
  \includegraphics[width=1\textwidth]{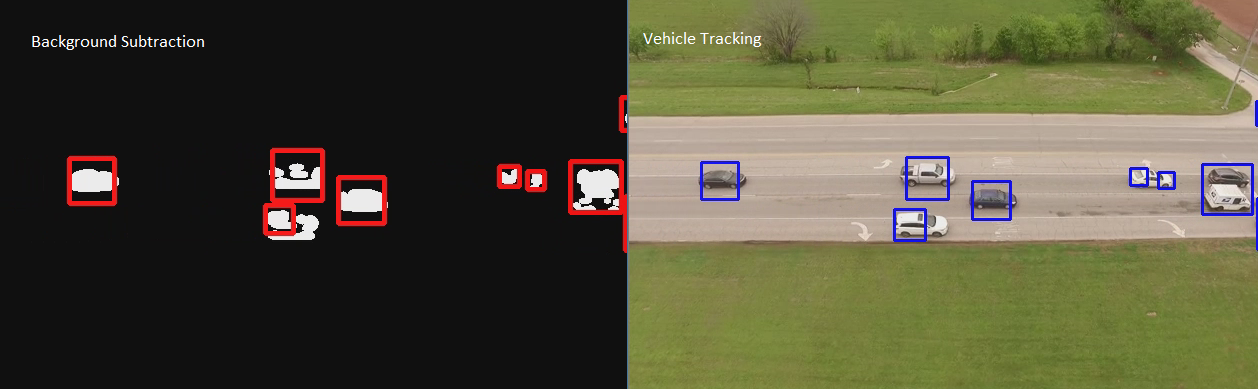}
   \caption{(Left) A sample background-subtracted image using MOG2. (Right) The detected blobs from the left.}
  \label{img:bgs_blobs}
\end{figure*}

An important step to background subtraction is the removal of small (noisy) movement by the background. This can occur due to wind blowing small objects, like plant life, or due to small instability in the camera feed. To remove these small perturbations, a Gaussian blur is applied to the image. The diameter of the blur is between 3 and 31 pixels, decided by the user through the user interface. An example of our background subtraction method can be seen in Fig. \ref{img:bgs_blobs}.

\subsection{Blob Detection}
Blob Detection \cite{blob_detection} is a method in computer vision that detects regions in a digital image that are different from the surrounding regions. The detector works by going through each pixel of the image and comparing it to its adjacent pixels. If the adjacent pixels are similar, by color in this case, then the method will group them together to form a ``blob". This blob object will have many attributes to it, with the most important being the center point of the blob and the size of the blob. The blob detection works together with the background subtraction to distinguish separate moving objects in the video frame. Since a moving object will be white in the background subtracted video and the background will be black, a filter can be put on the blob detector to make it find only the white blobs. 

Once again, the OpenCV library~\cite{opencv_library} was used in the program for its simple blob detector. We set up the blob detector to filter by color and area. The minimum blob size will be determined by the user using the graphical user interface, since the program needs to be able to detect a wide range of moving objects in different conditions. The blob detector is then used on the video with the background subtracted from it. If the detector is working, it will draw a blue box around the moving object and add it to a list of objects to track. An example of this is shown in Fig.~\ref{img:bgs_blobs}.

By examining Fig.~\ref{img:bgs_blobs}, the several instances can be found wherein the background subtraction has left gaps in the detected vehicles. These gaps can segment the blobs for a given vehicle, potentially reducing them to below the minimum blob size. To account for this, the output of the background subtraction is passed through a convolutional kernel which expands the size of each pixel from single pixel into a circle of diameter between 3 and 9, decided by the user.  
This improves detection of vehicles at the risk of increasing noise and merging distinct vehicles into a single blob. However, if applied sparingly and to a sufficiently noiseless background subtraction frame, the benefits vastly outweigh the cons.

The next thing the program will look at is whether the blob is a car or not. In order to determine if it is a car, the software will check the size of the blob, the speed of the blob, and how long the blob has been active. It makes sure the blob has been active for at least 8 frames before allowing it to be a car. If the blob passes the car test cases, then the box around the car will be green and the speed of the car in miles per hour will be shown right above the box. Out of these cars, the program will determine which cars will be considered to be in the queue. To be added to the queue list, a car must be going below a certain speed (25 miles per hour is currently being used) and going from left to right in the camera's field of view (FOV). Once the final queue list is created, the program compares the positions of the cars to each other to find the leftmost car which is declared as the end of the queue. A red box will be placed around the car and the GPS coordinates and speed will be displayed above the car and on the GUI.

\subsection{Car Location Calculation}
To calculate the physical location of a given pixel, telemetry data containing the heading, GPS location and the height of the UAV via lidar was used, along with the pixel location of objects detected in the recorded footage. Several approximations were made in order to decrease the development time for the GPS calculations (e.g., roll and pitch of the UAV were neglected, due to the high degree of stability of the DJI M600). These approximations were later determined to be sufficiently accurate during testing. The GPS coordinate calculation method is as follows.

\begin{figure}[htbp!]
  \centering
  \includegraphics[width=0.5\textwidth]{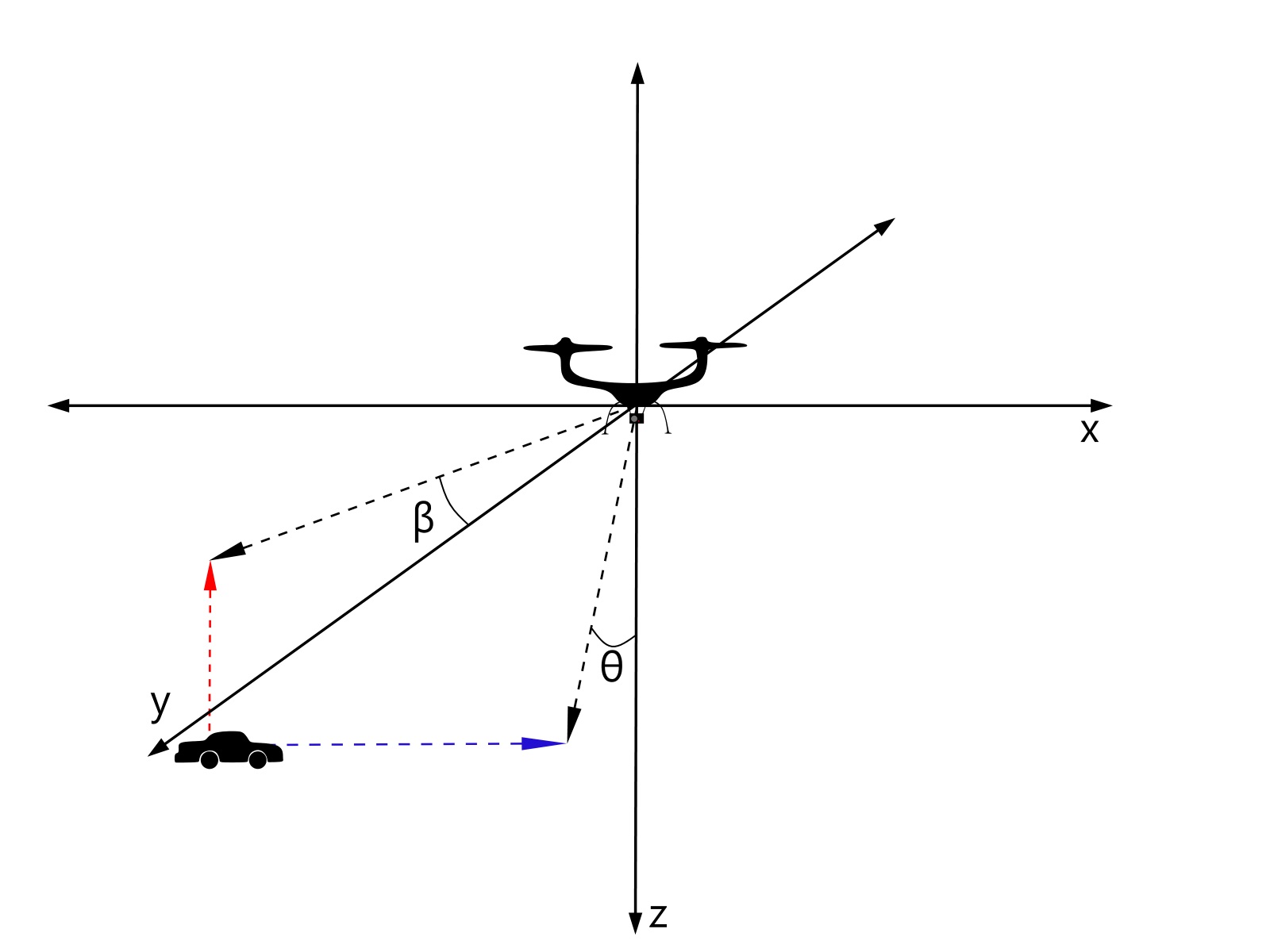}
  \caption{A representation of the drone in the Cartesian plane.}
  \label{img:test2}
\end{figure}

First, the coordinate system is defined as a Cartesian plane with the y-axis pointing directly ahead of the UAV, and the x-axis pointing directly to the side of the UAV. Naturally, the z-axis is defined as the vertical distance from the UAV, and this displacement is given from the lidar telemetry data. Lastly, $\theta$ is defined as the angle off the z-axis projected into the y-plane, and $\beta$ is defined as the angle off the y-axis projected onto the x-plane (Fig.~\ref{img:test2}).

To begin actual calculations, the y offset angle, $\alpha$, is found from the number of y pixels in a given row, the row index, and the focal length of the camera
\begin{equation}
    \alpha = \arctan\left(\frac{n_{y}-i_{row}}{2f}\right).
\end{equation}
Then, $\theta$ is found by adding  the center angle of the camera to the y offset angle, $\alpha$
\begin{equation}
    \theta = \alpha + \theta_{camera}.
\end{equation}
Using the height, $h$, given from the lidar, the y displacement, $\Delta_y$, from the UAV was found by utilizing the following trigonometric equation
\begin{equation}
    \Delta_y = h * \tan(\theta).
\end{equation}

Finding the x displacement, $\Delta_x$, was accomplished in a very similar fashion as the y displacement. First, the x offset angle, $\beta$, is found using the following equivalent to (1)
\begin{equation}
    \beta = \arctan\left (\frac{n_{x}-i_{column}}{2f}\right ).
\end{equation}
Since the center of the camera has no angle off the y-axis,
\begin{equation}
    \Delta_x = \Delta_y * \tan(\beta).
\end{equation}

The heading of the UAV, $\Theta$, is given from the telemetry data, which is used to rotate the displacement in the x and y directions into the more useful coordinate axes, North and East.
\begin{equation}
    \begin{bmatrix}
        \Delta_{North}\\
        \Delta_{East}
    \end{bmatrix}
=
    \begin{bmatrix}
        \sin(\Theta) && \cos(\Theta)\\
        \cos(\Theta) && \sin(\Theta)
    \end{bmatrix}
    \begin{bmatrix}
        \Delta_x\\
        \Delta_y
    \end{bmatrix}
\end{equation}

Finally, once the displacement in the North and East directions is known, the objects latitude, $\phi$, and longitude, $\lambda$, are found via the following equations,
\begin{equation}
    \phi_{object} = \frac{\Delta_{North}}{R_{Earth}} + \phi_{UAV}
\end{equation}
\begin{equation}
    \lambda_{object} = \frac{\Delta_{East}}{R_{Earth}*\cos(\phi_{object})} + \lambda_{UAV}
\end{equation}
with $R_{Earth}$ equaling the Earth's radius, $\phi_{UAV}$ equaling the UAV's measured latitude, and $\lambda_{UAV}$ equaling the UAV's measured longitude.

It was necessary to test whether the GPS coordinates assigned to the objects in the video were accurate, which requires accurate data from the UAV. A total of 12 videos were captured to test the GPS coordinates with different places, angles, and heights in each video. Once the footage and data were processed, a test function was written in the program that would print the pixel location and calculated GPS coordinate in the command prompt. For each video, it would find some key points (like trees, road signs, and road corners) throughout the video to test whether those coordinates matched up with their real coordinates. The test function would run the program, pick four or five points and then enter them in Google Maps to see where the points were located compared to their actual location. 
A simple program was then created to calculate how far the guessed points and actual points were away from each other using Haversine’s formula. Once the GPS coordinates were tested and determined to be accurate, testing could be done on the speed calculations. To do this, a vehicle was driven at a constant speed while being recorded by the UAV. Using the velocity equation with the Haversine function, the program calculated the distance between GPS coordinates and using the frame rate to determine how long it took to get from one point to another. This calculated velocity was compared to the known speed. This was used to calibrate the calculated speed. 

\begin{figure*}[htbp!]
  \centering
  \includegraphics[width=0.65\textwidth]{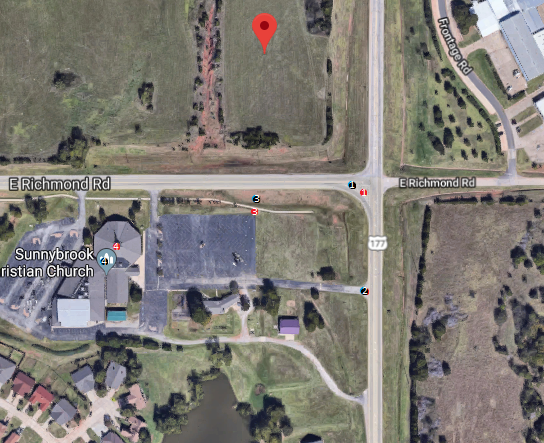}
  \caption{The results of our accuracy testing. The pin point represents the position of the UAV, the red dots represent the key point that was clicked on during testing, and the black dots represent the calculation that our program made.}
  \label{img:location_testing}
\end{figure*}

Fig.~\ref{img:location_testing} shows the accuracy of our program on Google Maps. The most accurate point in this test was point 2 with an error of 2.816\,m and the least accurate point was point 4 with an error of 26.33\,m. 

\subsection{Software Integration and Threading}
In order to separate the frequency of telemetry data sampling and frame-rate, as well as to handle the GUI without high latency, the software utilizes multithreading. Each thread handles a portion of the code in parallel with the others, allowing the operating system to assign multiple cores to the software. The threads are shown as the bottom three dotted-square regions in Fig.~\ref{img:Software_flow}.

The threads communicate by updating shared-memory variables. Most of these variables are read-only by one or more threads and write-only by one thread, so for the most part, thread-lockings have been avoided which helps prevent parallelism bottlenecks.

\section{Testing}
For testing of the software and hardware, most of the components and software parts were tested individually. These individual tests are outlined earlier in the paper under the Hardware and Software sections. The full image processing software was tested on multiple prerecorded videos taken from traffic cameras and UAVs. A full system test was performed at a traffic light to test all the working UAV hardware, real-time video, and image processing software together. A traffic light was selected as cars form a queue when the light is red. There are two directions that traffic flows, and the cars have to reduce speed significantly when approaching the light, which models high traffic queues very well.\footnote{DOT and FAA approval is needed to test in work zones. Testing was conducted at a traffic light, since permission to fly over a highway was not granted at the time of testing.} Below additional simplifying assumptions were made for testing real-time detection of EOQ of traffic and establishing it as a proof of concept: \label{testing}

\begin{itemize}
  \item Moderate weather conditions: the system does not need to operate in rain, snow, or windy weather conditions. UAV should be able to handle a moderate amount of wind (15\,mph).
  \item Ideal roads: the road is considered to be straight, and without intersections – typical of long stretches of highway for which the overall research is aimed.
  \item Stationary UAV footage: the UAV is assumed to be stationary while recording video.
  \item Minimum communication interference: this project will not need to account for loss-of-control situations beyond built-in systems of the UAV.
\end{itemize}

During the test, video footage was captured so that the software could process it as well as live transmission of the UAV state information was communicated back to the program. The video was post-processed with the image processing software and the end of the queue was calculated. Minor refinements were made to the code to account for the reduced speed of the cars approaching the stop light compared to what would be seen on a highway. The EOQ for our testing was set to be one-third of the posted speed limit. This speed was viewed as a significant enough speed reduction to be detected reliably. The best speed for defining the EOQ may vary depending on the situation.

\begin{figure*}[htbp!]
  \includegraphics[width=0.5\textwidth]{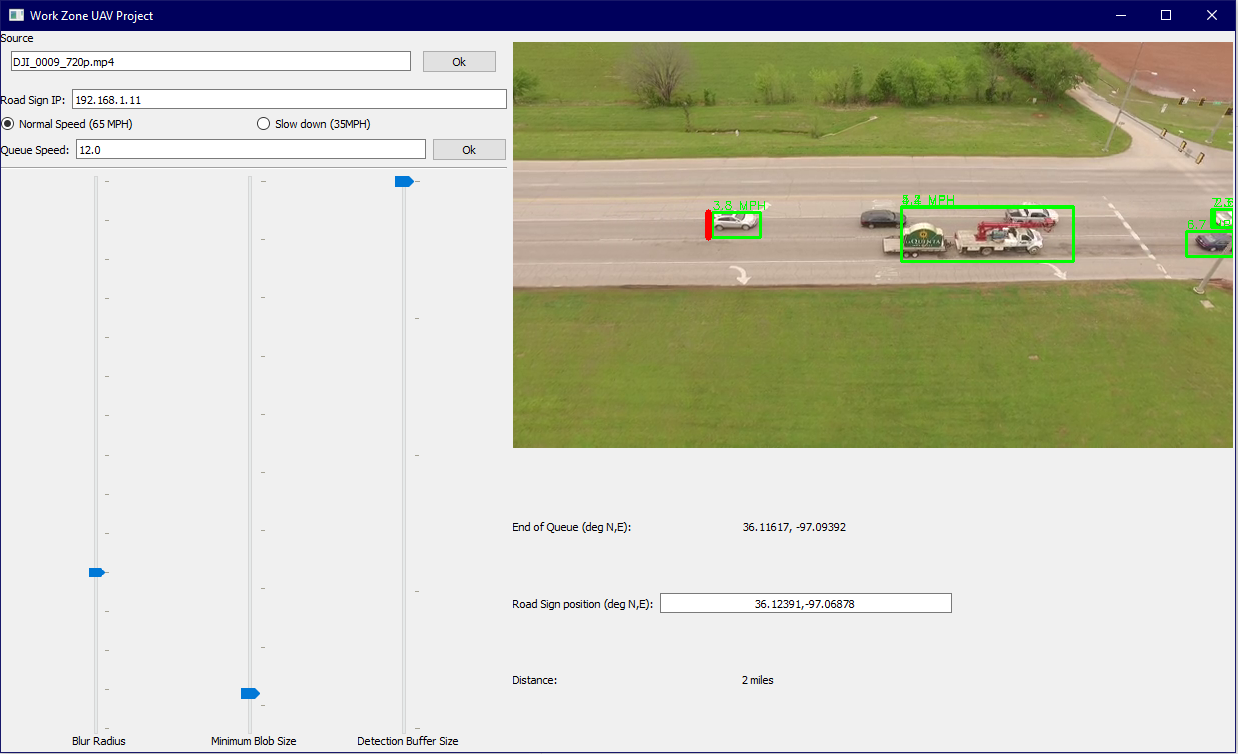}
  \includegraphics[width=0.5\textwidth]{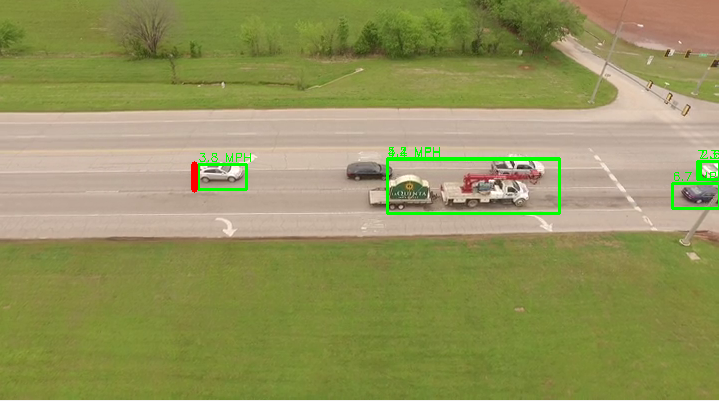}

  \caption{(Left) Program running EOQ detection.  (Right) Close up of video feed running in the program.}
  \label{img:EOQD}
\end{figure*}

A screenshot of the running program was taken and is shown in Fig.~\ref{img:EOQD}.  The red bar marks the EOQ location in the video feed.  Tracked vehicles are boxed in green.  The vehicles' speeds are posted on top of their boxes.

\section{Discussion}
Regulations on UAVs are one of the major limitations for deploying them to monitor traffic and find the EOQ. One human operator will be required to fly the UAV within the operator's unaided vision. The UAV used for testing has a flight time of 30 minutes. Two UAVs would need to be required for continuous traffic monitoring. How often UAVs will be flown will depend on the situation and requirements. For example, whenever there is highway maintenance/repair work or any accident that causes one or more lanes to be blocked, the authority will deploy UAVs to monitor the traffic situation and control the warning signs. UAVs will be deployed more frequently if they are used for general traffic monitoring at intersections. The UAV used in testing is able to monitor 858 ft length of straight road with its camera at an altitude of 400 ft and an FOV angle of 94 degrees.

The proposed approach of EOQ monitoring using image processing may encounter additional challenges during night-time or in adverse weather conditions, such as rain or snowfall. FAA regulations require anti-collision lighting to fly UAVs at night. However, simple image processing techniques can produce false objects due to car headlights and brake lights. Thermal imaging can be used at night to monitor the positions and speeds of moving vehicles without being affected by the headlights and brake lights of the vehicles. Additional pre-processing steps, often utilizing machine learning techniques, will be necessary before EOQ detection to eliminate artifacts caused by rain and snow in the video's image frames. Optimization and testing of these methods could be used to create a 24-hour traffic monitoring solution, effective in all weather conditions. However, these studies will be left as future works.


\section{Conclusions and Future Directions}
In conclusion, EOQ detection is possible in real time with video footage taken from UAVs.  As stated in the introduction, this EOQ information can be used to dynamically adjust signage for workzones, making roads safer and more efficient.  Video footage, and accurate positional data is all that is needed to find the location of the queue.  

The major next step that would be taken is to test the software on real traffic queues in work zones. Another future step would be to perform robust testing of the proposed method to create a quantitative analysis of any potential errors caused by the image processing or other factors. Current traffic models could then be compared to the data to find any discrepancies. This would reveal what improvements need to be made too. Also, it may be a good idea to implement a Convolutional Neural Network (CNN) to help improve car detection and make it easier to modify the project for other applications like counting car types going through an intersection and detecting wrong way drivers.

\section*{List of Abbreviations}

\begin{tabular}{ll}
ATC & Air Traffic Control \\
CCTV & Closed-circuit Television \\
CNN & Convolutional Neural Network \\
DOT & US Department of Transportation \\
EOQ & End-of-Queue \\
FAA & Federal Aviation Administration \\
FOV & Field of View \\
GPS & Global Positioning System \\
GPU & Graphics Processing Unit \\
GUI & Graphical User Interface \\
HDD & Hard Disk Drive \\
ITS & Intelligent Traffic System \\
MAV & Manned Aerial Vehicle \\
MOG & Mixture of Gaussian \\
OpenCV & Open Source Computer Vision Library \\
PCMS & Portable Changeable Message Sign \\
SDK & Software Development Kit \\
SSD & Solid-state Drive \\
STMC & State Traffic Management Center \\
UAV & Unmanned Aerial Vehicle \\
\end{tabular}

\section*{Conflict of Interest}
The authors declare that they have no conflict of interest.

\bibliographystyle{IEEEtran.bst}
\bibliography{references}

\begin{thebibliography}{10}
\providecommand{\url}[1]{#1}
\csname url@samestyle\endcsname
\providecommand{\newblock}{\relax}
\providecommand{\bibinfo}[2]{#2}
\providecommand{\BIBentrySTDinterwordspacing}{\spaceskip=0pt\relax}
\providecommand{\BIBentryALTinterwordstretchfactor}{4}
\providecommand{\BIBentryALTinterwordspacing}{\spaceskip=\fontdimen2\font plus
\BIBentryALTinterwordstretchfactor\fontdimen3\font minus \fontdimen4\font\relax}
\providecommand{\BIBforeignlanguage}[2]{{%
\expandafter\ifx\csname l@#1\endcsname\relax
\typeout{** WARNING: IEEEtran.bst: No hyphenation pattern has been}%
\typeout{** loaded for the language `#1'. Using the pattern for}%
\typeout{** the default language instead.}%
\else
\language=\csname l@#1\endcsname
\fi
#2}}
\providecommand{\BIBdecl}{\relax}
\BIBdecl

\bibitem{nhtsa_2016}
{National Highway Traffic Safety Administration}, \emph{Traffic Safety Facts 2017: A Compilation of Motor Vehicle Crash Data}.\hskip 1em plus 0.5em minus 0.4em\relax National Highway Traffic Safety Administration, National Center for Statistics and Analysis, U.S. Department of Transportation, 2019.

\bibitem{jazayeri_cai_zheng_tuceryan_2011}
A.~Jazayeri, H.~Cai, J.~Y. Zheng, and M.~Tuceryan, ``Vehicle detection and tracking in car video based on motion model,'' \emph{IEEE Transactions on Intelligent Transportation Systems}, vol.~12, no.~2, p. 583–595, 2011.

\bibitem{yang_hou_jhang_2013}
M.~T. Yang, J.~S. Hou, and R.~K. Jhang, ``Traffic flow estimation and vehicle-type classification using vision-based spatial–temporal profile analysis,'' \emph{IET Computer Vision}, vol.~7, no.~5, p. 394–404, 2013.

\bibitem{Barmpounakis_vlahogianni_golias_2016}
E.~N. Barmpounakis, E.~I. Vlahogianni, and J.~C. Golias, ``Unmanned aerial aircraft systems for transportation engineering: Current practice and future challenges,'' \emph{International Journal of Transportation Science and Technology}, vol.~5, no.~3, pp. 111--122, 2016.

\bibitem{Kanistrtas_Martins_Rutherford_2013}
K.~Kanistras, G.~Martins, M.~J. Rutherford, and K.~P. Valavanis, ``A survey of unmanned aerial vehicles ({UAV}s) for traffic monitoring,'' in \emph{2013 International Conference on Unmanned Aircraft Systems, ICUAS 2013 - Conference Proceedings}, 2013, pp. 221--234.

\bibitem{Wang_Liu_Zhou_2017}
X.~Wang, J.~Liu, and Q.~Zhou, ``Real-time multi-target localization from unmanned aerial vehicles,'' \emph{Sensors (Switzerland)}, vol.~17, no.~1, 2017.

\bibitem{multi_tracking_aerial}
G.~M{\'a}ttyus, C.~Benedek, and T.~Szir{\'a}nyi, ``Multi target tracking on aerial videos,'' in \emph{ISPRS Workshop on Modeling of Optical Airborne and Space Borne Sensors}.\hskip 1em plus 0.5em minus 0.4em\relax Istanbul, Turkey: IAPRS, 2010.

\bibitem{li_ye_chung_kolsch_wachs_bouman_2016}
J.~Li, D.~H. Ye, T.~Chung, M.~Kolsch, J.~Wachs, and C.~Bouman, ``Multi-target detection and tracking from a single camera in unmanned aerial vehicles (uavs),'' in \emph{2016 IEEE/RSJ International Conference on Intelligent Robots and Systems (IROS)}, 2016, pp. 4992--4997.

\bibitem{gleason_nefian_bouy_fong_bebis_2011}
J.~Gleason, A.~V. Nefian, X.~Bouyssounousse, T.~Fong, and G.~Bebis, ``Vehicle detection from aerial imagery,'' in \emph{2011 IEEE International Conference on Robotics and Automation}, 2011, pp. 2065--2070.

\bibitem{baykara_biyik_gul_onural_ozturk_yildiz_2017}
H.~C. Baykara, E.~Bıyık, G.~Gül, D.~Onural, A.~S. Öztürk, and I.~Yıldız, ``Real-time detection, tracking and classification of multiple moving objects in uav videos,'' in \emph{2017 IEEE 29th International Conference on Tools with Artificial Intelligence (ICTAI)}, 2017, pp. 945--950.

\bibitem{Collins_Barooah_Hespanha_2016}
G.~Collins, P.~Barooah, and J.~Hespanha, ``Geotrack: An autonomous closed-loop target tracking system for small {UAV} networks,'' \emph{International Conference and Exhibition of Unmanned Aerial Vehicles}, 2010.

\bibitem{Zhou_2010}
G.~Zhou, ``Geo-referencing of video flow from small low-cost civilian {UAV},'' \emph{International Journal of Transportation Science and Technology}, vol.~7, no.~1, p. 156–166, 2010.

\bibitem{khan_ectors_bellemans_janssens_wets_2017}
M.~A. Khan, W.~Ectors, T.~Bellemans, D.~Janssens, and G.~Wets, ``{UAV}-based traffic analysis: A universal guiding framework based on literature survey,'' \emph{Transportation Research Procedia}, vol.~22, pp. 541–--550, 2017.

\bibitem{wang_chen_yin_2016}
L.~Wang, F.~Chen, and H.~Yin, ``Detecting and tracking vehicles in traffic by unmanned aerial vehicles,'' \emph{Automation in Construction}, vol.~72, p. 294–308, 2016.

\bibitem{huang_chen_yang_cheng_2017}
C.~Huang, P.~Chen, X.~Yang, and K.~T.~T. Cheng, ``Redbee: A visual-inertial drone system for real-time moving object detection,'' \emph{2017 IEEE/RSJ International Conference on Intelligent Robots and Systems (IROS)}, 2017.

\bibitem{Salvo_Caruso_Scordo_2014}
G.~Salvo, L.~Caruso, and A.~Scordo, ``Urban traffic analysis through an {UAV},'' \emph{Procedia - Social and Behavioral Sciences}, vol.~28, pp. 1083--1091, 2014.

\bibitem{Hosseinpoor_Samadzadegan_Dadrasjavan_2016}
H.~R. Hosseinpoor, F.~Samadzadegan, and F.~DadrasJavan, ``Pricise target geolocation and tracking based on {UAV} video imagery,'' \emph{International Archives of the Photogrammetry, Remote Sensing and Spatial Information Sciences - ISPRS Archives}, vol.~41, pp. 243--249, July 2016.

\bibitem{Khan_Ectors_bellemans_Wets_2017}
M.~A. Khan, W.~Ectors, T.~Bellemans, D.~Janssens, and G.~Wets, ``Unmanned aerial vehicle-based traffic analysis: A methodological framework for automated multi-vehicle trajectory framework for automated multi-vehicle trajectory extraction,'' \emph{Transportation Research Record: Journal of the Transportation Research Board}, vol. 2626, no.~1, pp. 25--33, 2017.

\bibitem{Khan_Ectors_bellemans_Wets_2018}
------, ``Unmanned aerial vehicle-based traffic analysis: A case study for shockwave identification and flow parameters estimation at signalized intersections,'' \emph{Remote Sensing}, vol.~10, no.~3, 2018.

\bibitem{bruin_booysen_2015}
A.~D. Bruin and M.~J. Booysen, ``{Drone-Based traffic flow estimation and tracking using computer vision},'' in \emph{South African Transport Conference Project: Intelligent Transport Systems}, 2015, pp. 869--878.

\bibitem{Dille_Grocholsky_Singh_2011}
M.~Dille, B.~Grocholsky, and S.~Nuske, ``Persistent visual tracking and accurate geo-location of moving ground targets by small air vehicles,'' in \emph{Infotech@ Aerospace 2011}, 2011, p. 1558.

\bibitem{Hoose2016}
N.~{Hoose}, ``Queue detection using computer image processing,'' in \emph{Second International Conference on Road Traffic Monitoring, 1989.}, 1989, pp. 94--98.

\bibitem{Zanin2003}
M.~Zanin, S.~Messelodi, and C.~Modena, ``An efficient vehicle queue detection system based on image processing,'' in \emph{12th International Conference on Image Analysis and Processing, 2003.Proceedings.}, 2003, pp. 232--237.

\bibitem{Albiol2011}
A.~Albiol, A.~Albiol, and J.~M. Mossi, ``Video-based traffic queue length estimation,'' in \emph{2011 IEEE International Conference on Computer Vision Workshops (ICCV Workshops)}, 2011, pp. 1928--1932.

\bibitem{Cai2010}
Y.~Cai, W.~Zhang, and H.~Wang, ``Measurement of vehicle queue length based on video processing in intelligent traffic signal control system,'' in \emph{2010 International Conference on Measuring Technology and Mechatronics Automation}, vol.~2, 2010, pp. 615--618.

\bibitem{Fathy1995}
M.~Siyal and M.~Fathy, ``Real-time measurement of traffic queue parameters by using image processing techniques,'' in \emph{Fifth International Conference on Image Processing and its Applications, 1995.}, 1995, pp. 450--454.

\bibitem{AlOkaishi2019}
W.~{Al Okaishi}, A.~Zaarane, I.~Slimani, I.~Atouf, and M.~Benrabh, ``{Vehicular queue length measurement based on edge detection and vehicle feature extraction},'' \emph{Journal of Theoretical and Applied Information Technology}, vol.~97, no.~5, pp. 1595--1603, 2019.

\bibitem{pcms1}
\BIBentryALTinterwordspacing
{PCMS-1210 Pro Series G3}. Ver-Mac Inc. Accessed on: 10-12-2020. [Online]. Available: \url{https://www.ver-mac.com/en/products/series/serie/message-signs/product/portable-changeable-message-sign-pcms/1}
\BIBentrySTDinterwordspacing

\bibitem{pcms2}
\BIBentryALTinterwordspacing
{Traffic Message Board 3 Lines}. Transportation Supply LLC. Accessed on: 10-12-2020. [Online]. Available: \url{https://www.trans-supply.com/pg/163/traffic-message-board-3-lines}
\BIBentrySTDinterwordspacing

\bibitem{Shirazi2015}
M.~S. Shirazi and B.~Morris, ``Vision-based vehicle queue analysis at junctions,'' in \emph{2015 12th IEEE International Conference on Advanced Video and Signal Based Surveillance (AVSS)}, 2015, pp. 1--6.

\bibitem{Craig2017}
C.~M. Craig, J.~Achtemeier, N.~L. Morris, D.~Tian, and B.~Patzer, ``{In-Vehicle Work Zone Messages},'' Minnesota Department of Transportation, Tech. Rep. June, 2017.

\bibitem{droneregulations}
\BIBentryALTinterwordspacing
Electronic {C}ode of {F}ederal {R}egulations. Accessed on: 10-12-2020. [Online]. Available: \url{https://gov.ecfr.io/cgi-bin/ECFR}
\BIBentrySTDinterwordspacing

\bibitem{dronewebsite}
\BIBentryALTinterwordspacing
Matrice 600. Shenzhen DJI Sciences and Technologies Ltd. Accessed on: 10-12-2020. [Online]. Available: \url{https://www.dji.com/matrice600}
\BIBentrySTDinterwordspacing

\bibitem{lidarwebsite}
\BIBentryALTinterwordspacing
{LW}20/c (100 m). Lightwire. Accessed on: 10-12-2020. [Online]. Available: \url{https://lightwarelidar.com/products/lw20-c-100-m}
\BIBentrySTDinterwordspacing

\bibitem{rfd900website}
\BIBentryALTinterwordspacing
{RFD} 900+ modem. RFDesign Pty Ltd. Accessed on: 10-12-2020. [Online]. Available: \url{https://store.rfdesign.com.au/rfd-900p-modem/}
\BIBentrySTDinterwordspacing

\bibitem{MOGref}
P.~KaewTraKulPong and R.~Bowden, ``An improved adaptive background mixture model for real-time tracking with shadow detection,'' in \emph{Video-based surveillance systems}.\hskip 1em plus 0.5em minus 0.4em\relax Springer, 2002, pp. 135--144.

\bibitem{opencv}
\BIBentryALTinterwordspacing
``How to use background subtraction methods.'' [Online]. Available: \url{https://docs.opencv.org/3.4/d1/dc5/tutorial_background_subtraction.html}
\BIBentrySTDinterwordspacing

\bibitem{blob_detection}
T.~Lindebeegr, ``Scale selection properties of generalized scale-space interest point detectors,'' \emph{Journal of Mathematical Imaging and Vision}, 2013.

\bibitem{opencv_library}
G.~Bradski, ``{The OpenCV Library},'' \emph{Dr. Dobb's Journal of Software Tools}, 2000.

\end{thebibliography}

\vspace{7mm}
\parpic{ \includegraphics [width=1in,height=1.265in,clip] {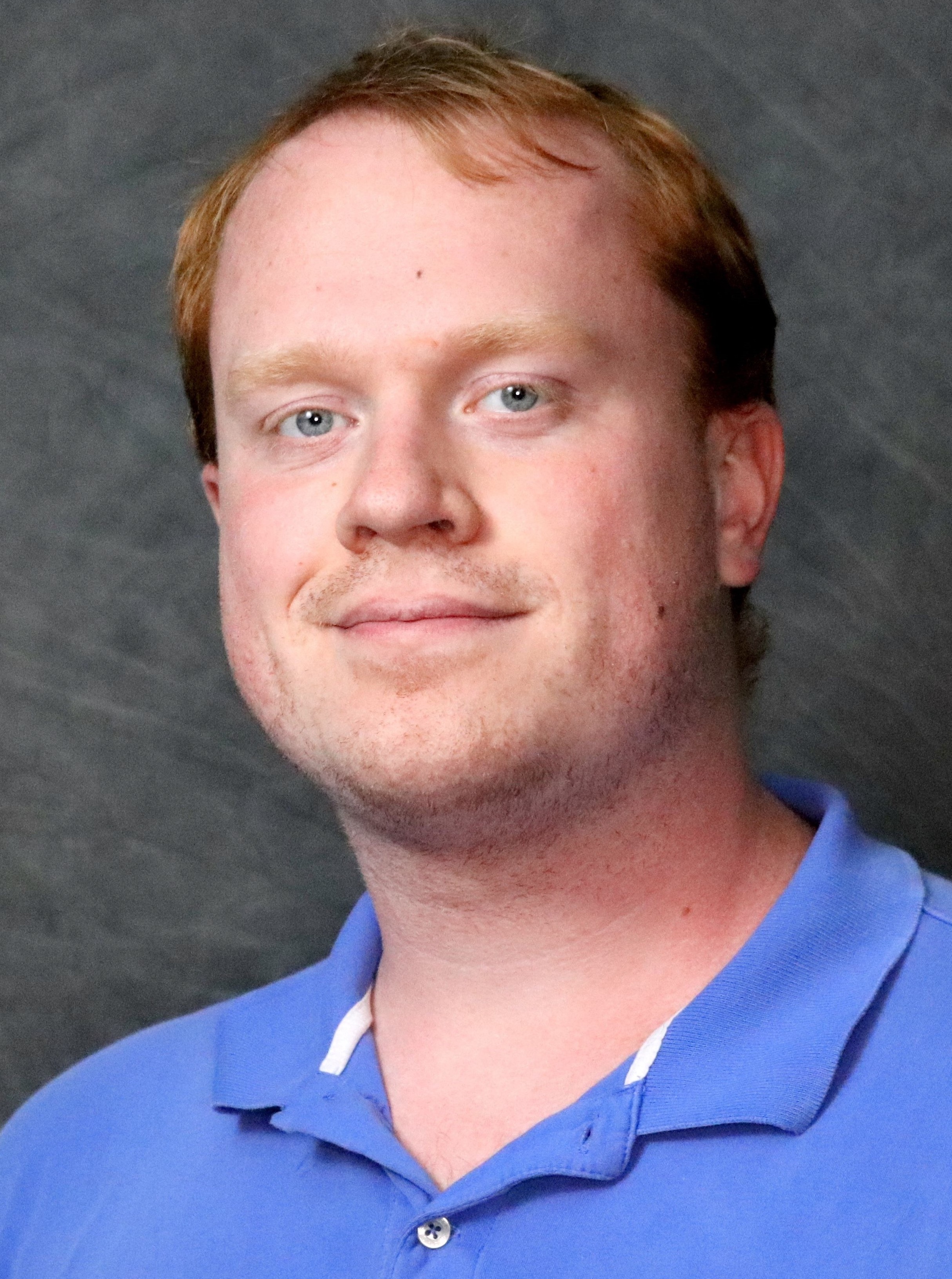}}

\noindent {\bf Russ Messenger} received his B.S. degree in electrical engineering from Oklahoma State University (OSU), Stillwater, OK, in 2019. Currently, he is pursuing his M.S. degree in electrical engineering under supervision of Dr. John F. O’Hara. He is working as a graduate research assistant at the Ultrafast Terahertz and Optoelectronic Laboratory, School of Electrical and Computer Engineering at OSU. His current research interests are in Internet of Things, terahertz band wireless communication, and terahertz time domain spectroscopy.
\vspace{3mm}

\parpic{ \includegraphics [width=1in,height=1.265in,clip] {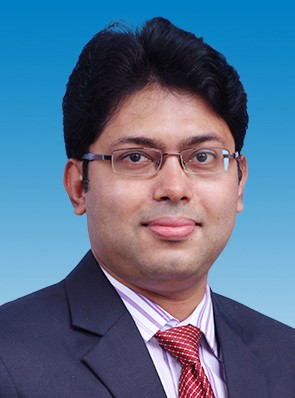}}

\noindent {\bf Md Zobaer Islam} received his B.Sc. degree in Electrical and Electronic Engineering in 2012 from Bangladesh University of Engineering and Technology, Dhaka, Bangladesh. He joined Oklahoma State University, Stillwater, OK as a Graduate Teaching Assistant to pursue his Ph.D. degree at the School of Electrical and Computer Engineering in Spring 2020 under supervision of professor Dr. Sabit Ekin. He has industry experience of 4 years at Bangladesh Telecommunications Company Ltd. in telecommunication and information technology (IT) sector and 3 years at Samsung R\&D Institute Bangladesh in software sector. His research interests include wireless light-wave sensing and machine learning.
\vspace{3mm}

\parpic{ \includegraphics[width=1in,height=1.265in,clip]{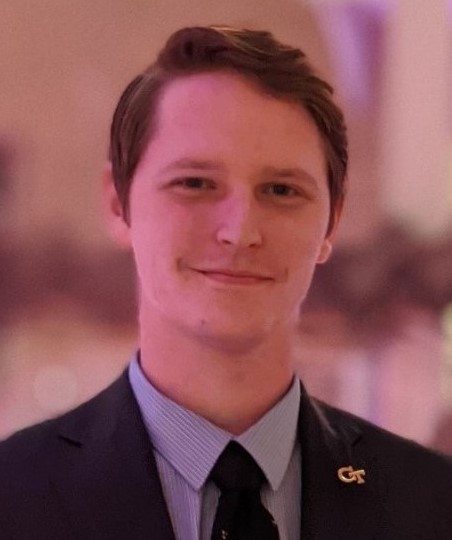}}\noindent {\bf Matthew Whitlock}  received his B.S. degree in Computer Engineering from Oklahoma State University (OSU), Stillwater, Oklahoma, USA in 2019. Currently, he is pursuing Ph.D. degree in Computer Science at Georgia Institute of Technology under supervision of Dr. Vivek Sarkar with research focus on hardware/software co-design for high performance computing.
\vspace{3mm}

\parpic{\includegraphics[width=1in,height=1.265in,clip]{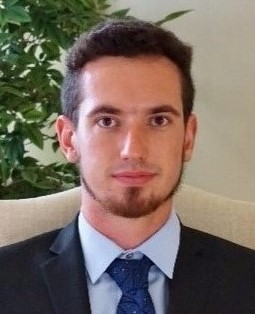}}
\noindent {\bf Erik Spong} received his B.S. degree in Mechanical Engineering from Oklahoma State University (OSU), Stillwater, Oklahoma, USA in 2019.

He has multiple summer and winter internship experiences during his undergrad studies. Currently, he is working at ExxonMobil as a marine integrity engineer.
\vspace{3mm}

\parpic{\includegraphics[width=1in,height=1in,clip]{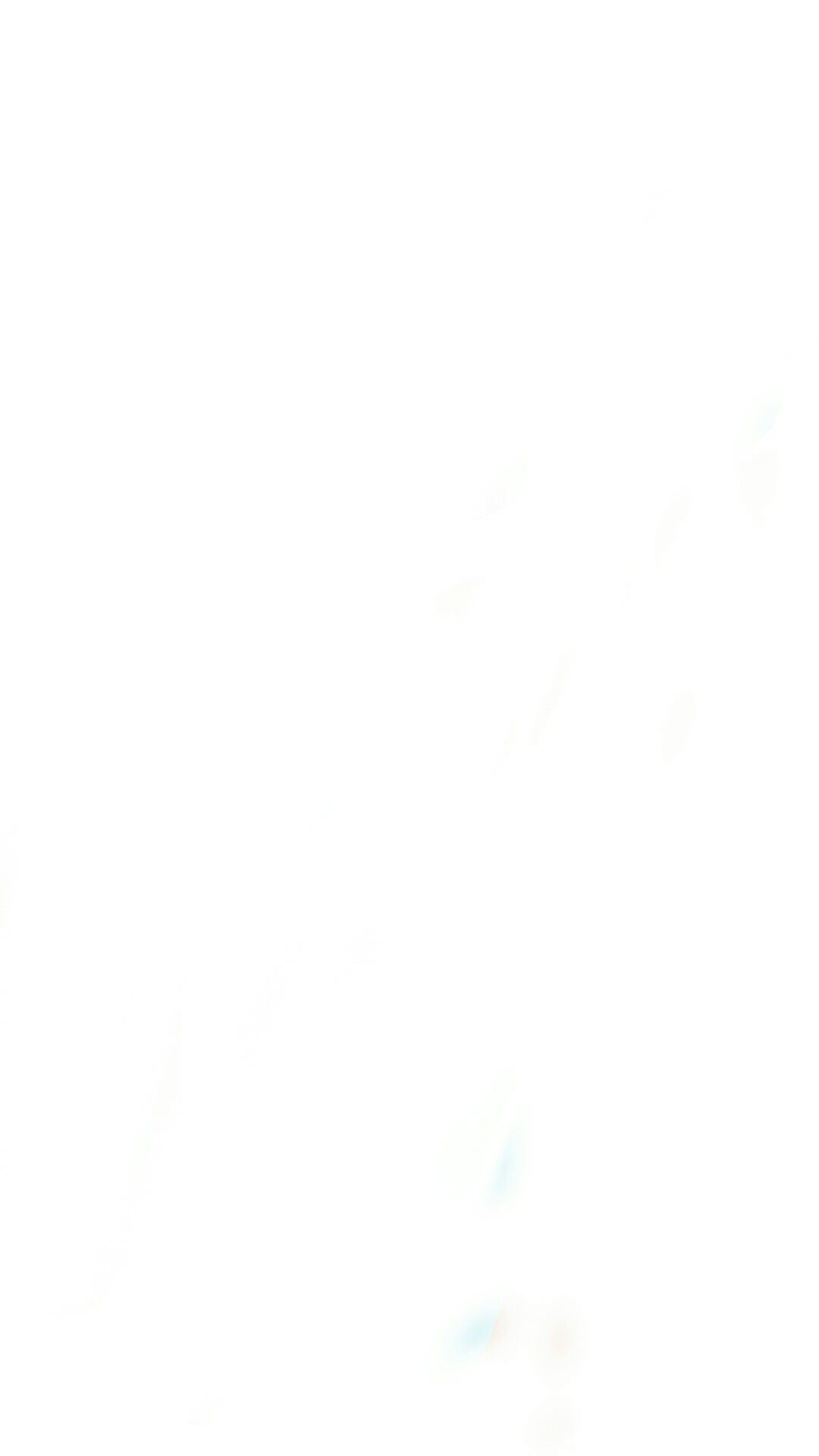}}
\noindent {\bf Nate Morton}  received his B.S. degree from Oklahoma State University (OSU), Stillwater, Oklahoma, USA in 2019. His major was computer engineering.
\vspace{3mm}

\parpic{\includegraphics[width=1in,height=1.265in,clip]{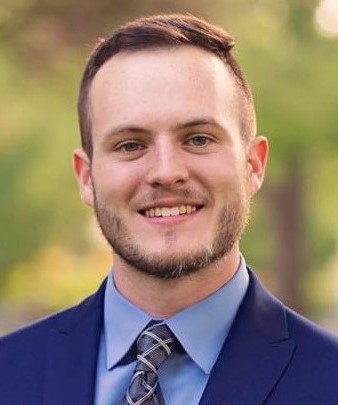}}
\noindent {\bf Layne Claggett} was born in Norman, Oklahoma, USA in 1997. He received the B.S. degree in mechanical engineering from Oklahoma State University (OSU), Stillwater, Oklahoma, USA in 2020.

Since 2019, He has been a systems engineer in the Defense, Space and Security sector of The Boeing Company, Inc.
\vspace{3mm}

\parpic{\includegraphics[width=1in,height=1in,clip]{white.jpg}}
\noindent {\bf Chris Matthews}  received his B.S. degree from Oklahoma State University (OSU), Stillwater, Oklahoma, USA in 2019. His major was computer engineering.
\vspace{3mm}


\parpic{\includegraphics[width=1in,height=1.265in,clip]{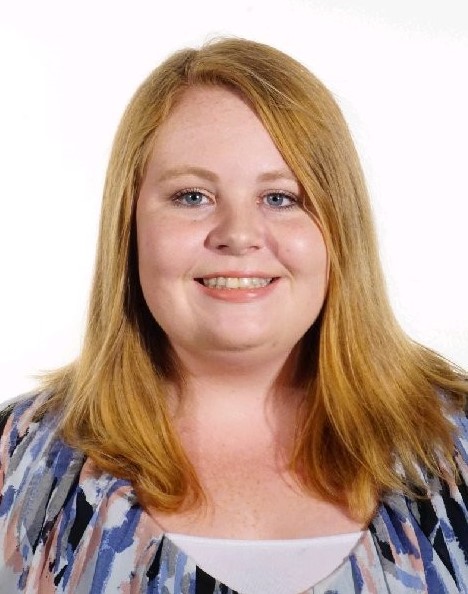}}
\noindent {\bf Jordan Fox} received his B.S. degree in Computer Engineering from Oklahoma State University (OSU), Stillwater, Oklahoma, USA in 2019.

Since 2019, she is working at TRU Simulation + Training as an associate software engineer.

\parpic{\includegraphics[width=1in,height=1.265in,clip]{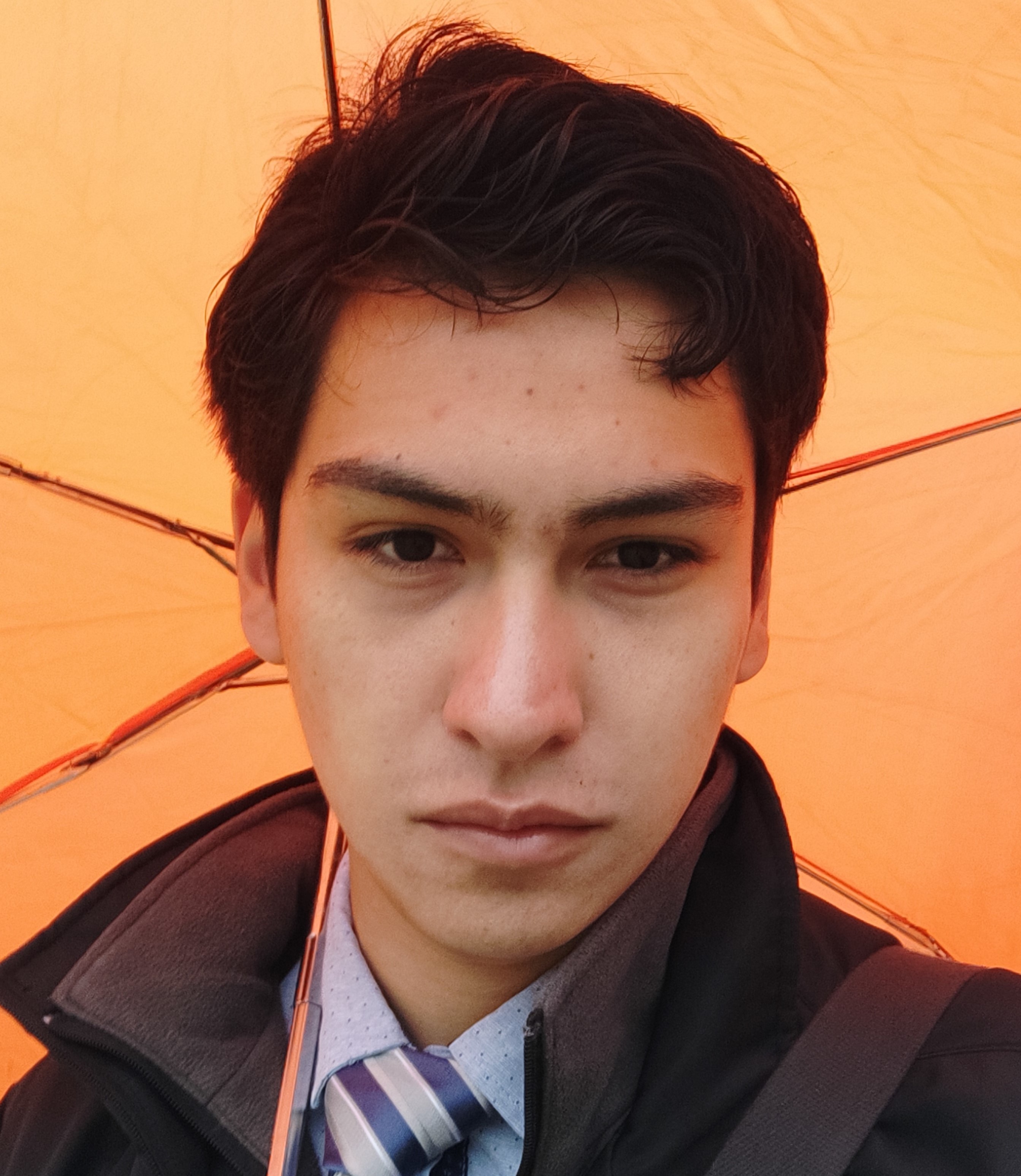}}
\noindent {\bf Leland Palmer}  received the B.S.E.E. degree in electrical engineering from Oklahoma State University, Stillwater, Oklahoma, USA in 2019.

He is currently a software engineer working for Ion Beam Applications.
\vspace{3mm}

\parpic{\includegraphics[width=1in,height=1.265in,clip]{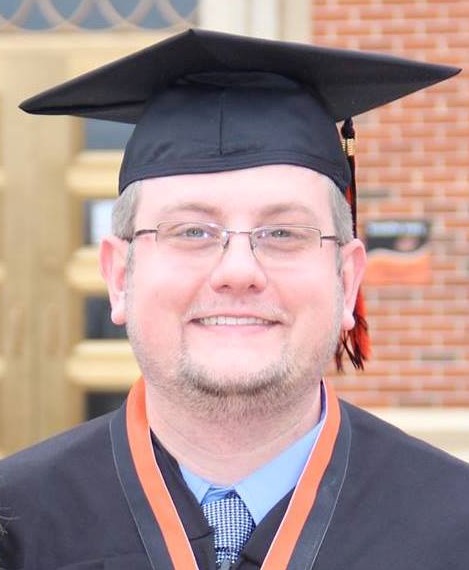}}
\noindent {\bf Dane C. Johnson} is a Research Engineer for the Unmanned Systems Research Institute at Oklahoma State University (OSU). He specializes in Computer and Electrical Engineering with a focus on autonomous systems and their applications. He has extensive experience in hardware and software development using commercial aviation grade sensors and was the project manager and lead developer for Oklahoma State University’s autopilot, Stabilis.  He also managed the development of General Electric’s methane detection drone, Raven.

\parpic{\includegraphics[width=1in,height=1.265in,clip]{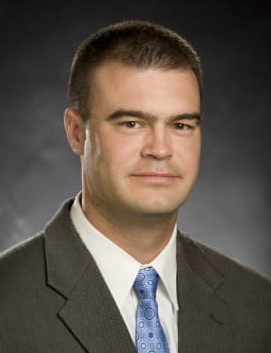}}
\noindent {\bf John F. O'Hara} received his BSEE degree from the University of Michigan in 1998 and his Ph.D. (electrical engineering) from Oklahoma State University (OSU) in 2003.  He was a Director of Central Intelligence Postdoctoral Fellow at Los Alamos National Laboratory (LANL) until 2006.  From 2006-2011 he was with the Center for Integrated Nanotechnologies (LANL) and worked on numerous metamaterial projects involving dynamic control over chirality, resonance frequency, polarization, and modulation of terahertz waves.  In 2011, he founded a consulting/research company, Wavetech, LLC specializing in automation and IoT devices.  In 2017 he joined OSU Electrical \& Computer Engineering, where he is currently an assistant professor and the Jack H. Graham Endowed Fellow of Engineering.  His current research involves terahertz wireless communications, terahertz sensing and imaging with metamaterials, IoT, and light-based sensing and communications.  He has 4 patents and around 100 publications in journals and conference proceedings.

\parpic{\includegraphics[width=1in,height=1.265in,clip]{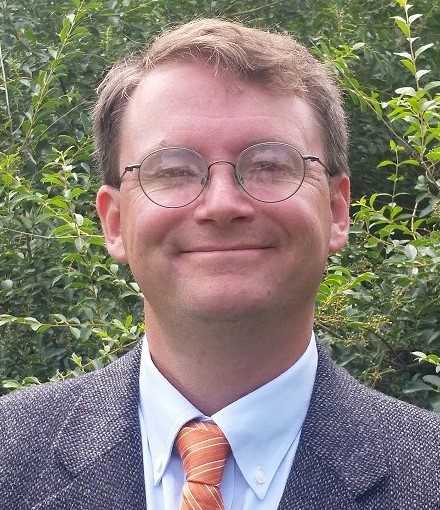}}
\noindent {\bf Christopher J. Crick} earned his BA in history at Harvard University in 1995 and his PhD in computer science at Yale University in 2009.  He was a postdoctoral research fellow at Brown University and has been a faculty member of Oklahoma State University since 2012, currently as Associate Professor.

Prof. Crick directs the Robotic Cognition Lab at OSU, where he focuses on robotics, artificial intelligence and machine learning.  The lab works to cultivate teamwork between humans and robots, as well as developing tools to understand, explain and trust the behavior and decision making of AI agents and machine learning systems. He has won grants from the National Science Foundation for projects from construction robots to weather research to big data analytics, and is the author of scores of papers.

\parpic{\includegraphics[width=1in,height=1.265in,clip]{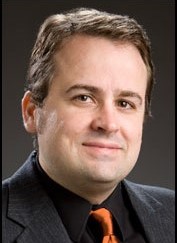}}
\noindent {\bf Jamey D. Jacob}  received the B.S. degree in aerospace engineering from the University of Oklahoma, in 1990, and the M.S. and Ph.D. degrees in mechanical engineering from the University of California at Berkeley, in 1992 and 1995, respectively. He has spent 10 years as a Professor with the Mechanical Engineering Department, University of Kentucky. He is currently the John Hendrix Chair and a Professor with the School of Mechanical and Aerospace Engineering and the Director of the Unmanned Systems Research Institute, Oklahoma State University. He is the author of over 100 papers and technical reports in the areas of unmanned systems, aerodynamics, flow control, and inflatable structures. He was a National Research Council Summer Faculty Fellow in the Air Force Research Laboratory at WPAFB, in 2003 and 2004, respectively.He currently serves on the Governor’s Aerospace and Autonomous Systems Council and as a President of the Unmanned Systems Alliance of Oklahoma.

\parpic{\includegraphics[width=1in,height=1.265in,clip]{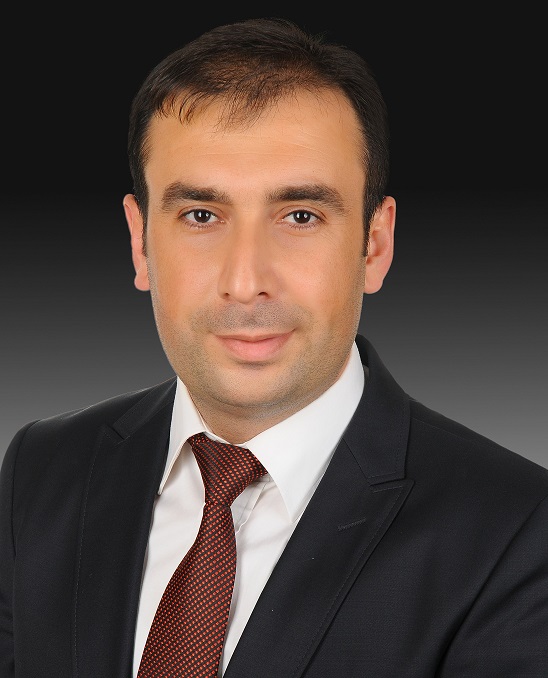}}
\noindent {\bf Sabit Ekin}  received his Ph.D. degree in Electrical and Computer Engineering from Texas A\&M University, College Station, TX, USA, in 2012. He has four years of industrial experience as a Senior Modem Systems Engineer at Qualcomm Inc., where he received numerous Qualstar awards for his achievements and contributions to cellular modem receiver design. He is currently an Associate Professor of Engineering Technology and Electrical \& Computer Engineering at Texas A\&M University. Prior to this, he was an Associate Professor of Electrical and Computer Engineering at Oklahoma State University. His research interests include the design and analysis of wireless systems, encompassing mmWave and terahertz communications from both theoretical and practical perspectives, visible light sensing, communications and applications, noncontact health monitoring, and Internet of Things applications.


\end{document}